\documentclass[a4paper, 10pt, conference]{cssconf}
\IEEEoverridecommandlockouts    
\usepackage{epsfig} 
\usepackage{amsmath} 
\usepackage{graphicx}
\usepackage{subcaption}
\usepackage[dvipsnames]{xcolor}

\usepackage{eso-pic}
\newcommand\AtPageUpperMyright[1]{\AtPageUpperLeft{
 \put(\LenToUnit{0.5\paperwidth},\LenToUnit{-1cm}){
     \parbox{0.5\textwidth}{\raggedleft\fontsize{9}{11}\selectfont #1}}
 }}
 
\newcommand{\conf}[1]{
\AddToShipoutPictureBG*{
\AtPageUpperMyright{#1}
}
}

\def\BibTeX{{\rm B\kern-.05em{\sc i\kern-.025em b}\kern-.08em
    T\kern-.1667em\lower.7ex\hbox{E}\kern-.125emX}}

\conf{International Conference on NeuroRehabilitation (ICNR2026), September 29-October 2 2026, Seoul, South Korea}

\title{\LARGE \bf Diffusion-Based Generation of Gait Trajectories}

\author{Damián Benasco, Juan Carballeira-Lopez,  Jaime Ramos-Rojas, Julio S. Lora-Millan, Antonio J. Del-Ama, \\David Rodriguez-Cianca, Pablo Lanillos*
\thanks{D.B., D.R-C and P.L are with the Cajal Neuroscience Center, Spanish National Research Council (CSIC), Madrid, Spain. J.C-L, J.R-R, J.S.L-M and A.J.D-A are with the Bioengineering Systems and Technologies Research Group, Rey Juan Carlos University (URJC), Madrid, Spain.}
\thanks{D.B. is funded by JAE-intro program from the CSIC. D.R-C is funded by the European Commission, NextGenerationEU, Momentum CSIC Programme. Work partially funded by DeepSelf project, Deutsche Forschungsgemeinschaft (nº 467045002). *Corresponding author: p.lanillos[at]csic.es.}
}

\begin{document}
\maketitle
\thispagestyle{empty}
\pagestyle{empty}

\begin{abstract}

Generation of musculoskeletal gait trajectories conditioned on patient-specific parameters remains a key challenge for wearable robotics and rehabilitation. Assistive systems such as lower-limb exoskeletons require reference trajectories that adapt to individual morphology and therapeutic goals while preserving biomechanical realism. Traditional approaches rely on hand-crafted gait templates or optimization procedures that scale poorly across subjects and walking conditions. In this work we explore conditional diffusion models for generating lower-limb joint-angle trajectories conditioned on gait parameters such as step length. We compare a baseline transformer diffusion model with a controllable diffusion transformer variant incorporating adaptive normalization and classifier-free guidance. Experiments on a dataset of 4,590 gait cycles show that diffusion models can generate realistic periodic gait trajectories while enabling some controllability variation in gait characteristics, highlighting their potential for personalized gait synthesis in assistive robotics.

\end{abstract}

\section{Introduction}
In wearable robotic systems, such as lower-limb exoskeletons, reference trajectories must adapt to patient morphology, pathology, and therapeutic goals while maintaining biomechanical plausibility~\cite{rigoli2024systematic}. Prior solutions range from hand-crafted gait templates to keypoint-based metaheuristics~\cite{carballeira2024metaheuristic}, but lack flexibility for continuous, conditioned generation. Recent advances in AI-driven methods, such as reinforcement learning~\cite{luo2024experiment,chavarrias2025adaptive,chavarrias_2025_15064551}, have showcased adaptive control for different morphologies. And generative modeling, e.g., diffusion models, have shown remarkable results in capturing complex data distributions in domains such as motion synthesis~\cite{tan2026gaitdynamics}, yet their application to clinical gait generation remains mostly unexplored. This work investigates diffusion-based models for generating lower-limb joint-angle trajectories conditioned on gait characteristics. Specifically, we evaluate (i)~a transformer-based model conditioned on biomechanical keypoints, and (ii)~a controllable Diffusion Transformer variant [5] incorporating physical gait parameters such as step length.
\begin{figure}[h]
\centering
\includegraphics[width=0.9\columnwidth]{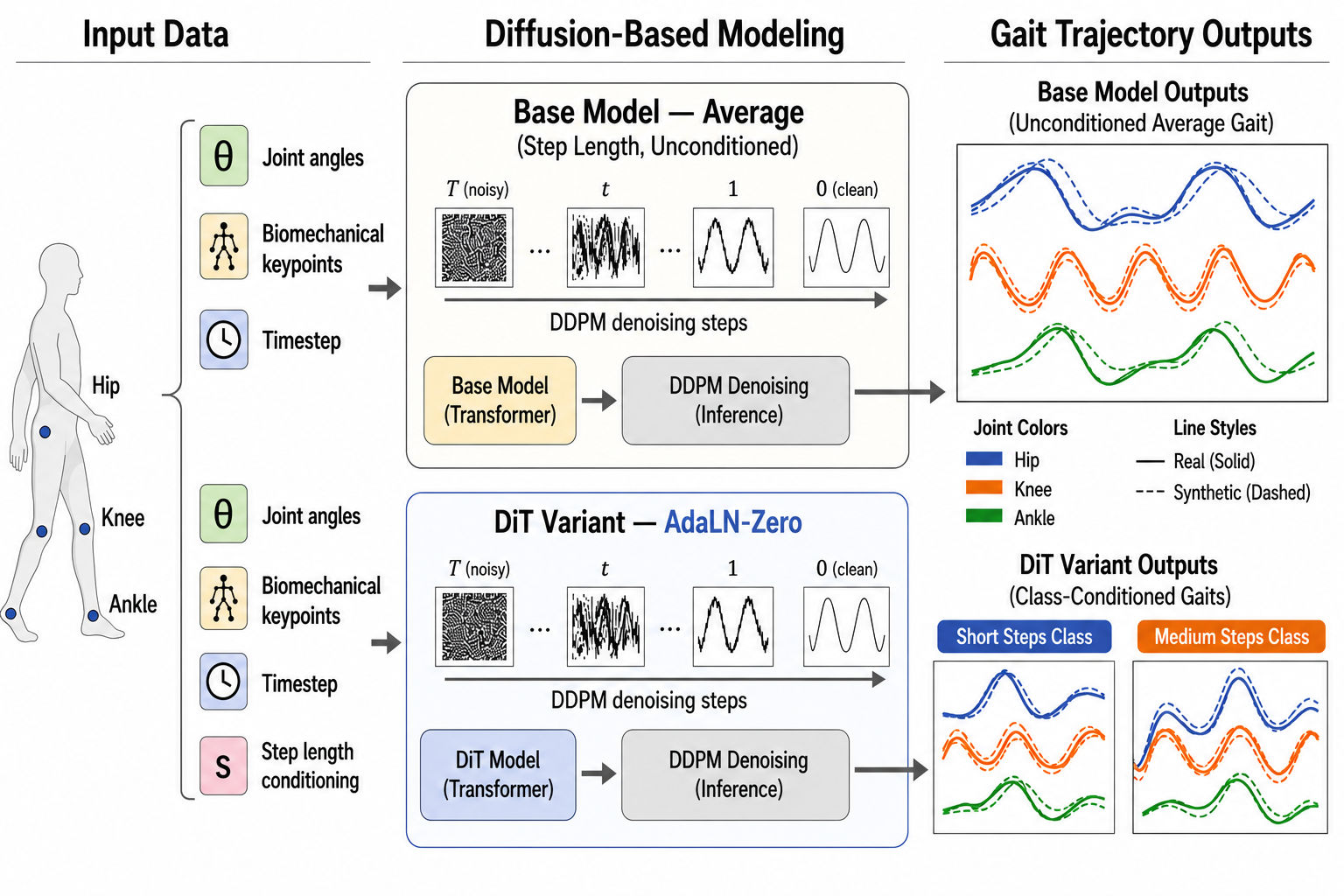}
 \caption{\textbf{Diffusion-based lower-limb gait generation}. Continuous kinematic trajectories generated by our diffusion model conditioned on biomechanical keypoints and step length via classifier-free guidance.}
\label{Hip_exo_model}
\end{figure}

\label{sec:methods:S1}
\section{Methods}
We frame gait synthesis as a denoising diffusion process. Our framework consists of a \textcolor{black}{baseline keypoint-conditioned Transformer diffusion model} and a Diffusion Transformer (DiT) extension with classifier-free guidance for step-length controllability.

\subsection{Dataset}

Two diffusion models were built using 4,590 gait cycles from 22 healthy subjects \textcolor{black}{(12 male, 10 female; age $26.5\pm7.7$ years, height $172.1\pm7.9$\,cm, weight $68.5\pm9.1$\,kg) instructed to walk at two distinct step lengths: short and medium. Each represented as 21 lower-limb joint angles (degrees) alongside biomechanical keypoints derived from VICON motion capture. To ensure strict class separation, step lengths were measured via forward kinematics and a 4 cm exclusion gap was applied, yielding 1,756 short steps (class 0, $< 48$ cm) and 2,834 medium steps (class 1,  $> 52$ cm).}

\subsection{Base Model: Keypoint-Conditioned Transformer}

Following MDM~\cite{tevet2023human}, the base model treats each frame as a token and conditions on biomechanical keypoint signals~\cite{koopman2014speed} (gait-phase pulses, event markers) via channel concatenation. It predicts the clean sample $\hat{x}_0 = f_\theta(x_t, t, c)$ with a linear noise schedule over 500 timesteps.

\subsection{DiT Variant: Controllable Generation via AdaLN-Zero}

To condition on physical gait parameters (e.g., step length), we replace layer normalization with AdaLN-Zero~\cite{peebles2023scalable}, where a vector composed of the timestep and gait-parameter embeddings produces per-layer shift, scale, and gate modulations. Kinematic pulse conditioning remains channel-concatenated at the input, while physical parameters are injected through AdaLN-Zero at every layer. Classifier-free guidance~\cite{ho2022classifierfree} is enabled via a learnable null embedding substituted during training ($p{=}0.25$).

\subsection{Training Objective}
The model minimizes $\mathcal{L} = \lambda_{\text{mse}} \mathcal{L}_{\text{mse}} + \lambda_{\text{cyc}} \mathcal{L}_{\text{cyc}} + \lambda_{\text{vel}} \mathcal{L}_{\text{vel}} + \lambda_{\text{fk}} \mathcal{L}_{\text{fk}}$, combining reconstruction, periodicity, smoothness, and a forward-kinematics foot-trajectory term (weights $\{2.0, 0.5, 0.1, 0.1\}$; $\lambda_{\text{fk}}$ ramped via curriculum learning).

\subsection{Model comparison}
Waveform fidelity is measured with the Pearson Correlation Coefficient ($R$) and Normalized RMSE (NRMSE). \textcolor{black}{Biomechanical validation uses per-joint Range of Motion (ROM) and bilateral Symmetry Index (left--right ROM agreement per joint). Cross-subject generalization is quantified via per-subject $R$.}
\section{Results and Discussion}
Table~\ref{tab:metrics} reports per-joint metrics for the base model (full dataset) and each DiT class evaluated against its corresponding real subset.

\begin{table}[h]
\centering
\caption{Sagittal-Plane Per-Joint Generation Metrics: Base Model vs.\ Conditioned DiT}
\label{tab:metrics}
\renewcommand{\arraystretch}{1.2}
\resizebox{\columnwidth}{!}{%
\begin{tabular}{l c c c c c c}
\hline
& \multicolumn{2}{c}{Base} & \multicolumn{2}{c}{DiT Class 0 (Short)} & \multicolumn{2}{c}{DiT Class 1 (Medium)} \\
\cline{2-3} \cline{4-5} \cline{6-7}
Joint & $R$ & NRMSE & $R$ & NRMSE & $R$ & NRMSE \\
\hline
L Hip  & 0.881 & 0.459 & 0.914 & 0.869 & 0.951 & 0.563 \\
L Knee  & 0.917 & 0.368 & 0.915 & 0.718 & 0.962 & 0.463 \\
L Ankle  & 0.540 & 1.510 & 0.628 & 2.045 & 0.613 & 1.676 \\
R Hip  & 0.879 & 0.463 & 0.900 & 0.898 & 0.952 & 0.555 \\
R Knee  & 0.903 & 0.403 & 0.918 & 0.718 & 0.956 & 0.480 \\
R Ankle  & 0.543 & 1.496 & 0.602 & 2.102 & 0.595 & 1.784 \\
\hline
\textbf{Average} & \textbf{0.777} & \textbf{0.783} & \textbf{0.813} & \textbf{1.225} & \textbf{0.838} & \textbf{0.920} \\
\hline
\multicolumn{7}{l}{\small Base: $57.4\pm3.7$\,cm\quad Class 0: $36.38\pm6.43$\,cm\quad Class 1: $68.58\pm6.95$\,cm} \\
\end{tabular}
}
\end{table}

\begin{figure}[h]
\centering
\includegraphics[width=\linewidth]{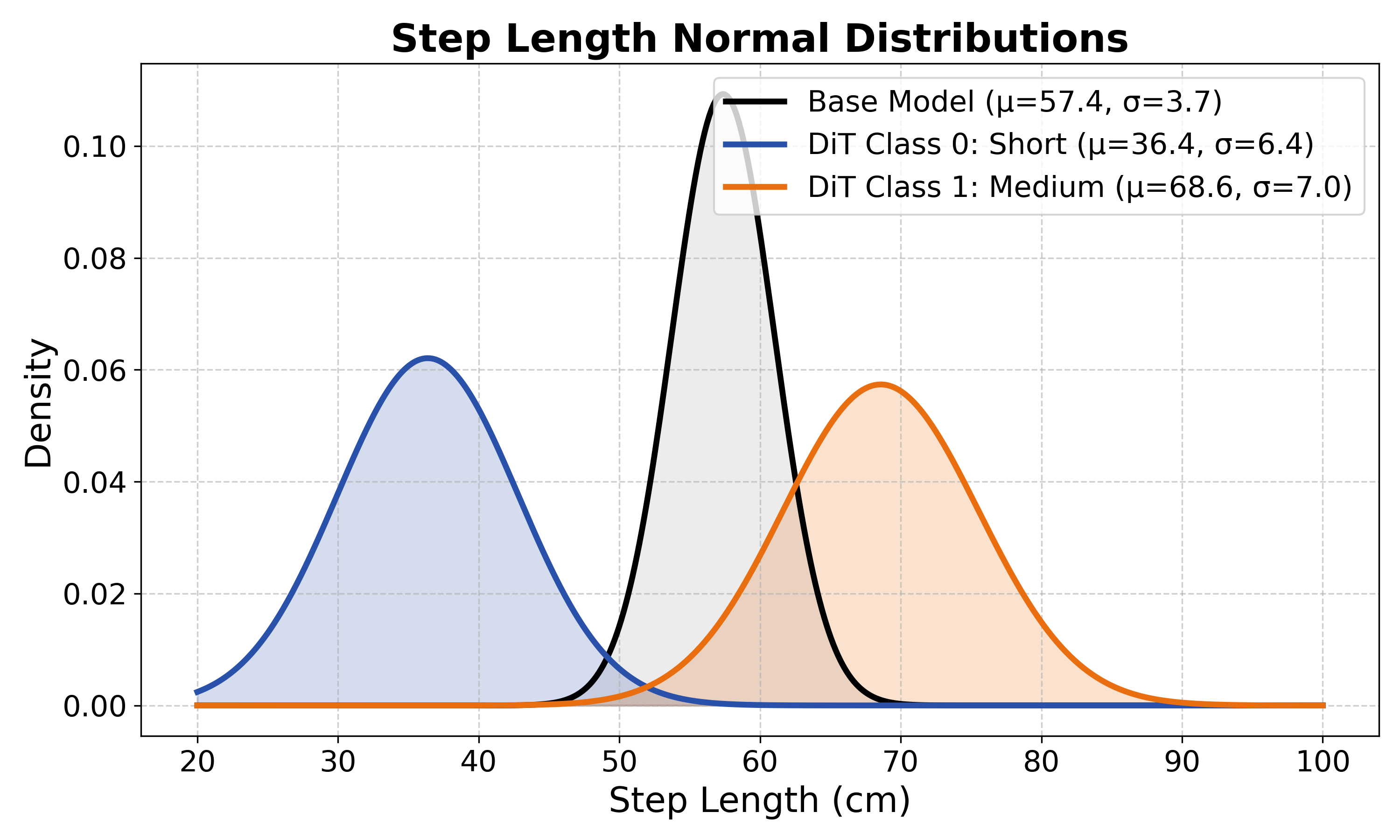}
 \caption{\textbf{Step-length conditioning}. Gait step length distribution of the generated data: base model vs step-length conditioned ().}
\label{fig:results}
\end{figure}

The base model captures gait dynamics ($R=0.777$) but converges to the dataset mean step length ($57.4{\pm}3.7$\,cm). The DiT achieves high fidelity for medium steps ($R=0.838$, NRMSE$=0.920$, $68.6{\pm}7.0$\,cm). Short-step conditioning shifts kinematics correctly ($36.4{\pm}6.4$\,cm, $R=0.813$) but yields higher error (NRMSE$=1.225$), \textcolor{black}{driven by greater inter-subject variability in short strides (knee ROM CoV\,$\approx$\,42\% vs.\ 26\%), causing MSE-trained diffusion to regress toward the mean. }Fig.~\ref{fig:results} confirms the DiT separates both classes from the base model's unimodal output.

\begin{figure}[h]
\centering
  \centering
  \includegraphics[width=\linewidth, height=200px]{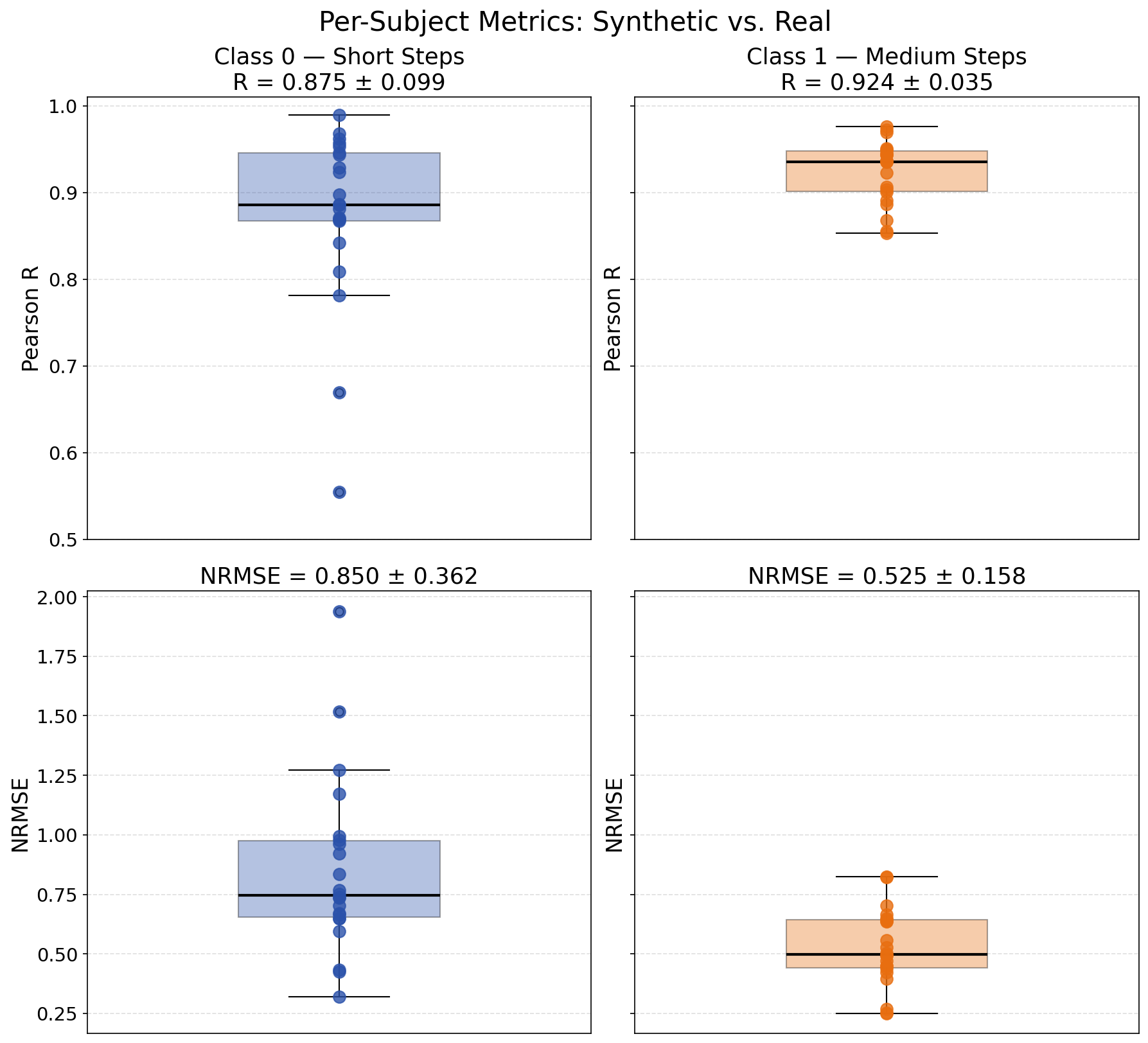}
\caption{\textbf{Subject specific validation.} Per-subject Pearson $R$ and NRMSE between real and synthetic across 22 subjects; short-step (blue, class~0) and medium-step (orange, class~1) gait.}
\label{fig:rom}
\hfill
\end{figure}

\textcolor{black}{Table~\ref{tab:biometrics} shows class~1 ROM error within $\pm6^{\circ}$; class~0 shows systematic underestimation (${\sim}10^{\circ}$ knee, ${\sim}6^{\circ}$ hip), consistent with MSE regression under high inter-subject variance. Symmetry is well preserved in class~1 but degrades in class~0 (hip/knee SI\,$\approx\,16$--$17\%$). Fig.~\ref{fig:rom} shows per-subject $R$ and NRMSE
across 22 subjects.}

\begin{table}[h]
\centering
\caption{Sagittal-plane ROM (deg): Real vs.\ Synthetic. Hip/Knee/Ankle averaged over left and right legs.}
\label{tab:biometrics}
{\footnotesize
\setlength{\tabcolsep}{4pt}
\begin{tabular}{l c c c c}
\hline
& \multicolumn{2}{c}{Class 0 (Short)} & \multicolumn{2}{c}{Class 1 (Medium)} \\
\cline{2-3} \cline{4-5}
Joint & Real & Synth & Real & Synth \\
\hline
Hip    & 29.9 & 24.1 & 52.4 & 57.2 \\
Knee   & 51.0 & 40.5 & 87.2 & 92.7 \\
Ankle  & 16.2 & 15.3 & 22.3 & 23.6 \\
\hline
\end{tabular}
}
\end{table}
\vspace{-0.1cm}
\section{Conclusions}

The proposed DiT generates biomechanically plausible, step-length-controllable gait trajectories \textcolor{black}{intended as kinematic references for adaptive exoskeleton controllers.} Class~1 achieves strong fidelity ($R=0.838$, ROM error $\leq\pm6^{\circ}$, well-preserved symmetry). Class~0 is harder due to distributional heterogeneity amplified by CFG null-embedding bias, producing amplitude underestimation and degraded symmetry. 
\textcolor{black}{While step length alone is insufficient to distinguish clinically relevant gait patterns (a slow shuffle and a fast short stride share similar step lengths but differ in cadence and speed), results show the potential of diffusion-based gait generation synthesis. Future work will address speed and subject-specific body dimensions conditioning and continuous multi-parameter conditioning for clinical deployment.}


\bibliographystyle{IEEEtran}
\bibliography{ICNR2026_diffusion}

\end{document}